\documentclass{article}
\usepackage[preprint]{neurips_2021}
\usepackage[utf8]{inputenc} % allow utf-8 input
\usepackage[T1]{fontenc}    % use 8-bit T1 fonts
\usepackage{hyperref}       % hyperlinks
\usepackage{url}            % simple URL typesetting
\usepackage{booktabs}       % professional-quality tables
\usepackage{amsfonts}       % blackboard math symbols
\usepackage{nicefrac}       % compact symbols for 1/2, etc.
\usepackage{microtype}      % microtypography
\usepackage{booktabs}
\usepackage[pdftex]{graphicx}
\usepackage{caption} 
\usepackage{amsmath}
\usepackage{csquotes}
\usepackage{enumitem}
\usepackage{tabulary}
\usepackage{longtable}
\usepackage{xcolor}

\title{MRCBert: A Machine Reading Comprehension Approach for Unsupervised Summarization}

\author{
Saurabh Jain \\
School of Computing \\
National University of Singapore \\
\texttt{saurabh@comp.nus.edu.sg} \\
\And
Guokai Tang\\
School of Computing \\
National University of Singapore \\
\texttt{guokai.tang@comp.nus.edu.sg} \\
\And
Lim Sze Chi \\
School of Computing \\
National University of Singapore \\
\texttt{sclim@comp.nus.edu.sg} \\
}

\begin{document}

\maketitle

\begin{abstract}
When making an online purchase, it becomes important for the customer to read the product reviews carefully and make a decision based on that. However, reviews can be lengthy, may contain repeated, or sometimes irrelevant information that does not help in decision making. In this paper, we introduce MRCBert, a novel unsupervised method to generate summaries from product reviews. We leverage Machine Reading Comprehension (\citet{mrc}), i.e. MRC, approach to extract relevant opinions and generate both rating-wise and aspect-wise summaries from reviews. Through MRCBert \footnote{Code: \url{https://github.com/saurabhhssaurabh/reviews_summarization.git}} we show that we can obtain reasonable performance using existing models and transfer learning, which can be useful for learning under limited or low resource scenarios. We demonstrated our results on reviews of a product from the Electronics category in the Amazon Reviews dataset. Our approach is unsupervised as it does not require any domain-specific dataset, such as the product review dataset, for training or fine-tuning. Instead, we have used SQuAD v1.1 dataset only to fine-tune BERT for the MRC task. Since MRCBert does not require a task-specific dataset, it can be easily adapted and used in other domains.
\end{abstract}

\section{Introduction}
With the rapid expansion of e-commerce platforms in recent years, online shopping has become relatively convenient and frequent attractive offers have made it even more popular amongst customers. One way of enabling shoppers to make informed decisions over their purchases is through the ratings and reviews made by previous customers. For the customer, it becomes important to only look at key aspects relevant to specific features of the product to make a decision quickly. However, it poses a challenge for the reader to scour through huge amounts of reviews which can be lengthy, contain repeated information about the product, or irrelevant information that does not give further insight to their decision making. This makes product-based opinion summarization in reviews important and useful for customers.

While opinion summarization has been widely done in supervised or semi-supervised contexts, modern deep learning methods rely on large amounts of annotated data or golden summaries that are not readily available in the opinion-summarization domain and are expensive to produce. Hence, the motivation of this paper is to conduct unsupervised summarization of user opinions relevant to a product feature. More specifically, we introduce MRCBert that uses existing open-sourced pre-trained models and demonstrate product-based opinion summarization using the machine reading comprehension (MRC) approach without using labeled or golden summaries.

Firstly, MRCBert performs a question answering task, in which it extracts relevant opinions from answers through product-centric questions. Then, the extracted opinions are input to an abstractive summarization model to produce aspect-wise summaries. To demonstrate proof of concept, we tested our approach on reviews from a single product using the publicly available Amazon electronics reviews dataset.

Our contributions can be summarised as follows:
\begin{enumerate}
    \item A novel unsupervised method that uses Machine Reading Comprehension approach to extract relevant opinions and generate both rating-wise and aspect-wise summaries from reviews.
    \item We have introduced a framework to generate summaries from reviews such that it can be easily scaled and used in any other domains. This framework does not require any domain-specific dataset. Our model can generate both aspect-wise and rating-wise summaries with the same pipeline, while other works have separate approaches to generate these.
    \item Our methodology can be used in limited or low resource scenarios. We have successfully demonstrated it with the help of a pre-trained MRC model and the concept of transfer learning.
    \item Introduced a unique sentiment accuracy metric to perform the unsupervised evaluation.
\end{enumerate}

\section{Related Works}
Earlier works on user reviews summarization like \citet{mcilory2015}, \citet{Joung2021} and \citet{Phong2015} used extractive techniques for topic modelling and keyword extraction. They are also often generically referred to as opinion mining techniques. Other approaches also attempted to solve the extractive opinion summarization task by finding attribute-value pairs that are of interest using aspect-based feature mining (\citet{Bafna2013}), or studying part-of-speech tagging to identify product features and their corresponding opinion words with the help of a sentiment classification module (\citet{4782570}).

However, these earlier works usually require golden summaries, or are rule-based methods that may not generalize well with noisy reviews datasets like the Amazon Reviews dataset, where there can be multiple topics unrelated to the product, misspelled words, colloquialisms, or misused sentence syntaxes. Later works that do unsupervised abstractive opinion summarization have pushed boundaries. As far as we are aware, the only two earlier methods dealing with unsupervised multi-document opinion summarization are MeanSum and Copycat. 

In the work of \citet{DBLP:journals/corr/abs-1810-05739}, the MeanSum model consists of two main components: (1) an auto-encoder module that learns representations for each review and constrains the generated summaries to be in the language domain and (2) a summarization module that learns to generate summaries that are semantically similar to each of the input documents. The authors have done this by training end-to-end with an autoencoder reconstruction loss to achieve both objectives. To obtain the opinion summary of one target product, the authors used the mean of all encoded reviews as the combined review representation and feed that into the decoder network to produce one summary.

Unlike aspect-based summarization in the work of \citet{Liu2012} that rewards the diversity of opinions, Copycat aims to generate summaries that represent a consensus i.e. dominant opinions in reviews. This is a relevant adaptation of the problem at hand as dominant reviews are useful for quick decision-making to get the overall feel of a product. The Copycat abstractive summarizer is similar to MeanSum with an encoder and decoder module that takes the mean of latent representations of reviews at inference time. However, their proposed architecture differs by using a hierarchical variational autoencoder (VAE) to study the latent representation of each review. Its decoder network also directly accesses all other reviews within a product, enabling it to capture common information across different reviews and retaining information of stronger relevance to the product. This is done by using a pointer-generator network that studies the distribution of words in each review and in each product combined. The authors showed superior performance in ROUGE R1, R2, and RL scores as compared to MeanSum. More details on ROUGE scores by \citet{lin-2004-rouge} are presented in Section \ref{section: evaluation metrics}.
\label{section: related works}

\section{Proposed Method}
\label{section: proposed method}
We fine-tuned BERT (\citet{devlin2018bert}) and DistilBERT (\citet{distilbert}) models with SQuAD v1.1 (\citet{squadv1}) dataset. Afterward, we extracted opinions from reviews, using a machine reading comprehension approach, which was used later to generate summaries. We have explained our approach in detail in the following sections.

\subsection{BERT}
In recent years, BERT has been one of the most successful deep learning architectures for NLP tasks. Although embedding layers can learn the contextual representation of texts after training on large-scale corpora, training a wide variety of networks for different tasks on limited datasets is unable to learn representations effectively. Unlike ELMO (\citet{elmo}) that learns additional text features by training a specific architecture for a specific task, BERT is intended for fine-tuning approach with a general model for all tasks. It is also desirable that an intelligent agent should learn more from data and use minimum prior human knowledge. 

Mainly there are two types of BERT architecture:
\begin{enumerate}
    \item BERT base: It contains 12 layers, 768 hidden dimensions, and 12 attention heads. It has total 110M parameters.
    \item BERT large: It has 24 layers, 1024 hidden dimensions, and 24 attention heads. It has total 340M parameters.
\end{enumerate}

\subsection{DistilBERT}
DistilBERT is a faster and lighter variant of BERT with some reduction in performance. According to the authors, it is 40\% lighter and 60\% faster than BERT, while it retains 97\% of its language understanding capabilities. We used it in our work to compare its performance with BERT on review summarization tasks when very little computation power is available.

\subsection{Fine-tuning BERT and DistilBERT}
We used SQuAD v1.1 (\citet{squadv1}) dataset to fine-tune BERT and DistilBERT. We used HuggingFace library\footnote{HuggingFace Library: \url{https://huggingface.co/}} for fine-tuning. SQuAD v1.1 contains more than $100,000$ pairs of questions and answers. We formulate fine-tuning task as follows. Given a question $q = (q_1, \dots, q_m)$ and a paragraph $p = (p_1, \dots, p_n)$, we formulate input as $x = [CLS],q_1,\dots,q_m,[SEP],p_1, ,\dots,p_n,[SEP]$, where [CLS] is a dummy token and [SEP] is a separator to separate question from paragraph. Let BERT(.) be the pre-trained BERT/DistilBERT model. We first get hidden representation of input x by passing it to pre-trained BERT/DistilBERT model as $h = BERT(x) \in R^{r_h * |x|}$, where $|x|$ is length of input sequence and $r_{h}$ is size of hidden dimension. Then hidden representation is passed to two separate dense layers (\citet{dense_layer}) followed by Softmax (\citet{softmax}) function: $l_1 = Softmax(W_1*h + b_1)$ and $l_2 = Softmax(W_2*h + b_2)$, where $W_1, W_2 \in R^{r_h}$ and $b1, b2 \in R$. Softmax is applied along the dimension of the sequence. Outputs $s$ and $e$ are span indices in paragraph $p$, computed from $l_1$ and $l_2$ as: 

\begin{equation}{
s = arg \hspace{0.5em} max_{index[SEP] < s < e}(l_1)
}\end{equation}
\begin{equation}{
e = arg \hspace{0.5em} max_{index[SEP] < e < |x|}(l_2)
}\end{equation}
Hence, final answer $a$ will always be a valid text span from $p_{s}$ to $p_{e}$. We used average cross entropy (\citet{cross_entropy_loss}) loss for both the pointers as our training loss:
\begin{equation}{
L = \frac{-\sum(log(l_1)OH(s)) - \sum(log(l_2)OH(e))}{2}
}\end{equation}
where $OH$ is representing one hot vector of ground truths $s$ and $e$.

\subsection{Opinion Extraction}
\label{section: proposed method, opinion}
We treated the opinion extraction task as a Machine Reading Comprehension task in which we considered a review as a paragraph and opinion on a feature as a question. At first, we manually extracted features from product descriptions. Since we did narrow down our scope to Computers, Tablets, and Laptops, all products have almost similar features. Therefore, manual extraction did not take much time. Alternatively, there are several research works available to extract features from product descriptions. Due to a lack of resources, we avoided using these research works. We used 2 question variants, both are targeted to ask about a product aspect. They are of the form \emph{"How is [feature]?"} and \emph{"What is opinion on [feature]?"}. We worked with a total of 10 features. These are Display, Memory, Speaker, Sound, Processor, WiFi, Battery, Brand, Operating System, and Camera.

We send a list of questions and a review to our fine-tuned BERT/DistilBERT model. Every question is represented as $q = (q_1,\dots, q_n)$ and review as $r = (r_1,\dots,r_m)$. The output of the model returns a list of indices pairs representing answer span $s$ and $e$ for each question, where $s$ represents start index and e represents end index. Then, text phrase from the reviews corresponding to these indices is considered as opinion on the respective feature i.e. $o = (r_s,\dots, r_e$. Then, we pass the list of reviews and opinions to our summarization module.

\subsection{Summarization Model}
As Copycat has claimed superior performance over MeanSum as an abstractive opinion summarization model, we experimented with pre-trained Copycat as a summarization module for our extracted opinions.\footnote{Copycat: \url{https://github.com/abrazinskas/Copycat-abstractive-opinion-summarizer}.} For more robust testing of MRCBert, we further experiment using a summarization model that is not tailored for opinion summarization. The objective is to investigate if MRCBert can perform with summarization models that are used for out-of-domain datasets, therefore further testing its scalability to other problem domains. As the unsupervised summarization models found thus far are both summarising opinions from reviews, the next resort is to use a supervised summarization model for this investigation. We used the DistilBART-12-6-cnn Summarizer available on HuggingFace. \footnote{DistilBART Summarizer API: \url{https://huggingface.co/sshleifer/distilbart-cnn-12-6}.}

Copycat has been introduced in Section \ref{section: related works}. In more detail, Copycat trains a Variational AutoEncoder to learn two latent representations: $z_i$ the latent representation of a review $r_i$ and a product latent representation $c$. During summary generation, Copycat infers the mean representation of a product given all $N$ reviews of the product (Eq. \ref{eqn:Copycat mean representation of a product.}), and then infer the mean representation of the review (Eq. \ref{eqn:Copycat mean representation of the review.}). To generate a summarizing review, the conditional language model is used to sample a review as in Eq. \ref{eqn:Copycat summary generation.}. $p_\theta (r|z^*, r_{1:N})$ in Eq. \ref{eqn:Copycat summary generation.} is done with the help of a pointer-generator network that maintains two-word distributions: one that assigns probabilities to words being generated using a fixed vocabulary, and another captures probabilities of words to be copied directly from other reviews.

\begin{equation}{
\label{eqn:Copycat mean representation of a product.}
c^* = \mathbb{E}_{c\sim q_\phi (c|r_{1:N})}[c]
}\end{equation}

\begin{equation}{
\label{eqn:Copycat mean representation of the review.}
z^* = \mathbb{E}_{z\sim p_\theta (z|c^*)}[z]
}\end{equation}

\begin{equation}{
\label{eqn:Copycat summary generation.}
r^* = \arg\max p_\theta (r|z^*, r_{1:N})[z]
}\end{equation}

The DistilBART-12-6-cnn has been trained on CNN/Dailymail (\citet{DBLP:journals/corr/SeeLM17}) and Extreme Summarization (XSum) datasets (\citet{Narayan2018DontGM}). These datasets contain document-summary pairs from news articles of various topics and are a different summarization problem domain from MRCBert, where the latter focuses on opinion summarization. Therefore, the DistilBART-12-6-cnn is suitable for our use case of investigating the scalability of MRCBert using pre-trained models from other problem domains. 

DistilBART-12-6-cnn originates as one of the applications of the Bart model introduced in the work of \citet{DBLP:journals/corr/abs-1910-13461}. It uses a standard seq2seq architecture with a bidirectional encoder and a left-to-right autoregressive decoder and can be fine-tuned for multiple downstream tasks such as the summarization task applied in MRCBert.

Besides using Copycat and DistilBART-12-6-cnn as the chosen summarization models to compare with, MRCBert uses a combination of the two. MRCBert uses the generated reviews by Copycat, which captures majority sentiment and handles misspelled words, colloquialisms, or misused sentence syntaxes with the idea that majority reviews are written in proper sentences. Then using the generated reviews, MRCBert taps on DistilBART-12-6-cnn to summarize the generated reviews. Section \ref{section: results} covers a comparison of the 2 summarization models when used separately and MRCBert, which uses a fusion of the two.

\subsection{MRCBert Summarization Framework}
We put together the opinion extraction module using MRC and the summarization module to generate rating-wise and aspect-wise summaries into a novel framework called MRCBert (Figure 1).
    \begin{figure}[!htb]
    \begin{center}
        \includegraphics[scale=0.4]{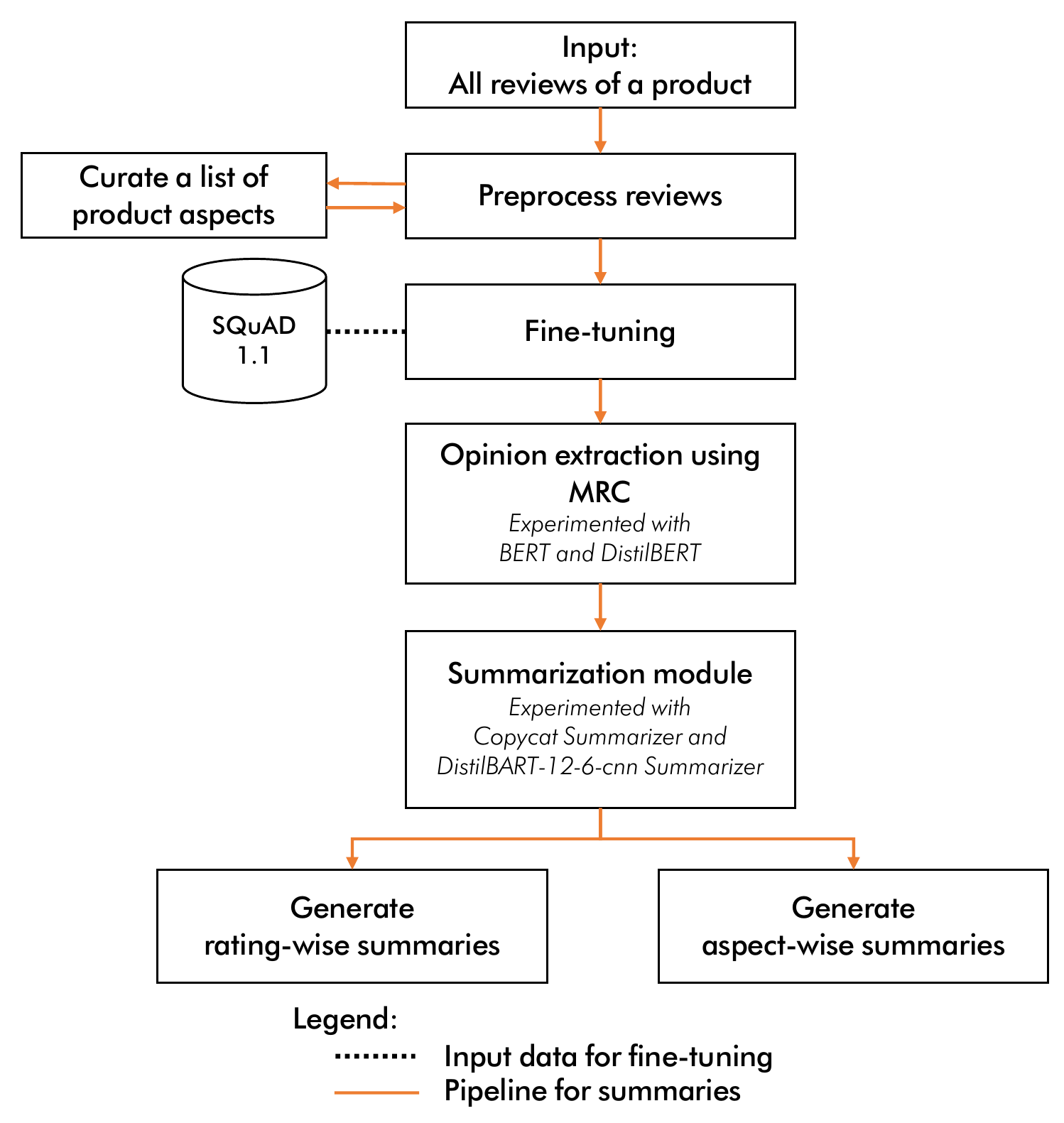}
        \caption{The proposed MRCBert summarization framework.}
    \end{center}
    \vspace{-1.5em}
    \end{figure}

\section{Evaluation Metrics}
\label{section: evaluation metrics}
To evaluate our generated summaries, we use the following set of metrics to evaluate their effectiveness. Given a set of summaries that consists of tens of thousands of summaries generated by our pipeline, evaluating them all manually would be labor-intensive if not impossible. Therefore we will use some metrics where each of these metrics can automatically evaluate some aspects of the summaries. 

\subsection{Sentiment Accuracy Metric}
We trained a sentiment prediction model using BERT \citet{devlin2018bert} as a preprocessing layer and as an encoding layer, we attached an output dense layer of size (None,6) that predicts the following: given a text that contains either review or summary, output the sentiment ranked from 0 to 5, which corresponds to the sentiment scale, for product reviews, the scale corresponds to ratings that user gave when they submitted their review text. We make the weights of both BERT layers trainable so that they can capture review-related linguistic knowledge during training. In total the trainable parameters of our model is 81,135,366 parameters. We trained the model on a subset of Amazon Reviews Dataset’s Electronic Category from the work of \citet{mcauley2015imagebased} that consists of 300,000 reviews for 20 epochs using an initial learning rate of $5\times10^{-5}$, batch size of 128 and ADAMW \citet{loshchilov2019decoupled} as optimizer. The results of our trained model are shown in table \ref{tab:sentiment-model-result}

\begin{table}[htb]
\centering
\caption{ Sentiment model result on test set}
\begin{tabular}{@{}lll@{}}
\toprule
 & Categorical Accuracy & Categorical Loss \\ \midrule
Sentiment Prediction Model & 0.746 & 0.674 \\ \bottomrule
\end{tabular}
\label{tab:sentiment-model-result}
\end{table}

To evaluate our approach, we score our model using the difference between predicted sentiment and an average rating of the reviews being summarized using equation \ref{eqn:sentiment_score}. We denote the score as $S_{sentiment}$. 

\begin{center}
\begin{equation}{
\label{eqn:sentiment_score}
S_{sentiment} = 1 - \frac {1} {N}  \sum_{i=1}^{N} \left[ \log_6 \left( \left | \frac{1}{k} \sum_{j=1}^{k} R(x_{j} ) - S(s_{i}) \right | + 1 \right) \right ]
 }\end{equation}
\end{center}
Where $S(s_i)$ is the sentiment score computed from the $i^{th}$ summary, 
$R(x_{j} )$ is the rating given $j^{th}$ review, 
$k$ is the total number of reviews being used to generate summary and $N$ is the total number of summaries, $log_6$ is a notation trick we use to make the score fall between $(0,1)$.

\subsection{ROUGE Scores}
Not only that we want to make sure the sentiment in our summary and the average sentiment in the reviews that the summary is generated from is similar. We also want to ensure that our summary contains similar information that is found in the set of reviews that the summary comes from. For example, if the reviews all mentioned screen size, we want to ensure that our summary also captures some summarized information about screen size. For this purpose, we use ROUGE-1 score to measure the overlap of unigram (each word) between the reviews and summary, and ROUGE-2 score to measure the overlap of bigram (two words) between the review and reference summary. 

To evaluate our approach, we score our model using the unigram and bigram difference between generated summary and the reviews being summarized using equation \ref{eqn:rouge_score}. We denote the score as $S_{Rouge}$. 

\begin{center}\begin{equation}{
\label{eqn:rouge_score}
S_{Rouge} = \frac {1} {N}  \sum_{i=1}^{N} \left[ \frac{1}{k}  \sum_{j=1}^{k} ROUGE_{i}(s_{i}, x_{j}) ) \right]
 }\end{equation}
\end{center}

Where $k$ is the total number of reviews being used to generate summary, $N$ is the total number of summaries and $ ROUGE_{i}(s_{i}, x_{j}) $ is the ROUGE-I score generated between $j^{th}$ review and $i^{th}$ summary.

\section{Experimental Setup}
\label{section: experimental setup}
\subsection{Dataset}
We used SQuAD v1.1 (\citet{squadv1}) and Amazon Reviews Data(2018) (\citet{amazon_reviews_2018}) in our work. We used SQuAD to fine tune BERT (\citet{devlin2018bert}) and DistilBERT (\citet{distilbert}) architectures. SQuAD dataset contains more than 100,000 pairs of question-answers extracted from more than 500 Wikipedia articles. For the summarization task, we have used the Amazon Reviews dataset. This dataset contains 233.1 million reviews collected in the range of May 1996 to October 2018. These reviews belong to products from 29 categories. Due to lack of resources, we focused only on the Electronics category. This category contains reviews of electronic gadgets and their accessories. Again, we did narrow down our scope and focused on reviews belonging to Computers, Tablets, and Laptops. In total, there were 131256 reviews in these categories.

\subsection{Data Preparation}
For fine-tuning BERT and DistilBERT, we have used the same SQuAD dataset as provided by the authors without data pre-processing. However, when applied to the Amazon Reviews dataset for the opinion extraction task, data pre-processing was done. 

As an extension to our analysis of MRCBert's performance, we tested MRCBert with and without our data pre-processing module. This is not including those pre-processing steps of the pre-trained models. When conducted without pre-processing, raw data from Amazon reviews were used as inputs to extract opinions. When conducted with pre-processing, we removed stop words, symbols, and numbers, and did lemmatization.

Although reviews have been provided as JSON objects, few reviews contain HTML tags and overlapped values of two or more keys. These issues created problems in parsing JSON objects. Therefore, we have removed such reviews in our work and dropped duplicated reviews (i.e. exact match in reviews and ratings). Finally, we have in total 4886 reviews from a single product.

\subsection{Experimental Details}
We executed our experiments on Tesla T4 16GB GPU machine with CUDA 11.2 installed. The BERT model for fine-tuning has 768 hidden sizes, 12 attention heads, and 12 hidden layers. We fine-tuned BERT for 3 epochs, it took around 3.85 hours to fine-tune. It had in total 87714 samples and per second 1.581 samples were processed. Maximum CPU memory usage was 0.17GB and maximum GPU memory usage was 9GB. We achieved an F1 score of 96.67 on the validation dataset and an F1 score of 88.60 on the test dataset.

The DistilBERT model has hidden size 768, 12 attention heads, and 6 hidden layers. It was fine-tuned for 3 epochs and it took 4.16 hours to fine-tune. It had total 87714 samples and per second 1.464 samples were processed. Maximum CPU memory usage was 0.17GB and maximum GPU memory usage was 4.57GB. We achieved an F1 score of 85.67 on the test dataset.

One limitation of the Copycat is that it currently supports only review summarization for up to 8 reviews. Hence, when applied to our dataset comprising of 4886 reviews, Copycat outputs multiple summaries, each summary corresponds to a group of up to 8 input reviews. As we are looking to have a single summary either product-wise (1 summary), ratings-wise (5 summaries per rating on the scale of 5), and aspects-wise (10 summaries for 10 aspects), we need a methodology to post-process and condense the multiple summaries into a single paragraph. 

To do so, we split the multiple summaries into multiple sentences. Each sentence will have its associated sentence embedding (\citet{reimers-2019-sentence-bert}). All sentence embeddings then go through agglomerative clustering with a distance threshold of 1.5 to keep the length of the post-processed summary moderate. We then sample the maximum length sentence in each cluster and join them to form a single summary. The maximum length sentence was chosen instead of the minimum length, as we find that the minimum length sentence could at times capture little or no information.

This post-processing is not needed when using the DistilBART-12-6-cnn Summarizer which can take in variable-length inputs. Hence, we joined all input sentences or extracted opinions to form a single document input to DistilBART-12-6-cnn Summarizer and it outputs a single summary.

\section{Results}
\label{section: results}
\subsection{Without additional pre-processing}
This section presents the evaluation results when no additional data pre-processing was conducted. Note that this is excluding the data pre-processing modules in the pre-trained models that were used.

\subsubsection{Rating-wise Reviews}
We evaluated the rating-wise summaries with and without the opinion extraction module to see how well the generated summaries reflect the input review ratings through the use of sentiment accuracy scores. When the summarization model used is Copycat, $S_{sentiment}$ is higher on average when given the actual reviews (without opinion extraction) as compared to using extracted opinions. However, when the summarization model used is DistilBART-12-6-cnn, $S_{sentiment}$ on average is the same. Hence, there is no clear distinction in Table \ref{tab: rating wise sent} that shows whether sentiment scores are better with or without opinion extraction.

A characteristic of rating-wise summaries that could have led to inconclusive results in Table \ref{tab: rating wise sent} may be the presence of mixed sentiments in the reviews. For example, there is a review rated 2 out of 5 but also has some positive experience. An excerpt of this review says: \emph{"Though the iPad is a fun toy to have around, the apps for it are pretty fun, and it has internet browsing capabilities, it only has about 80\% of the utility of using a computer.".} As a result, the generated summary by DistilBART-12-6-cnn with opinions extracted has captured some positive sentiment: \emph{"The camera takes lousy, grainy photos. (...) The iPad is a fun toy to have around."}. 

On the other hand, the rating 2 generated summary by DistilBART-12-6-cnn without opinion extraction captures only negative opinions. It says, \emph{"The iPad is extremely slow for a 32 gb (...) The software is outdated and the home bottom is broken."}. As a result of mixed sentiments present in reviews, the $S_{sentiment}$ scores for rating 2 summaries using DistilBART-12-6-cnn with opinion generated is poorer compared to without. We hypothesize likely a similar trend for other ratings.
\begin{table}[ht]
\centering
% \begin{center}
\caption{Sentiment accuracy scores for rating-wise reviews with and without opinion extraction, using Copycat and DistilBART-12-6-cnn summarization models}
\begin{tabular}{p{0.8cm}p{2.2cm}p{2.2cm}p{2.2cm}p{2.2cm}}
\toprule
{} & \multicolumn{2}{c}{Copycat} & \multicolumn{2}{c}{DistilBART-12-6-cnn} \\
\toprule Rating & $S_{sentiment}$ without opinion extraction & $S_{sentiment}$ with opinion extraction & $S_{sentiment}$ without opinion extraction & $S_{sentiment}$ with opinion extraction \\ \midrule
1  & 1.000   & 1.000     & 1.000     & 1.000  \\
2 & 0.226   & 0.226    & 1.000   & 0.613   \\
3   & 0.613   & 0.387   & 0.387 & 0.387  \\
4   & 1.000   & 1.000   & 0.613   & 0.613    \\
5  & 1.000 & 0.613   & 0.613  & 1.000     \\
\bottomrule
\end{tabular}
\label{tab: rating wise sent}
% \end{center}
\end{table}

\subsubsection{Aspect-wise Reviews}
\begin{figure}[h!]
\begin{center}
    \includegraphics[scale=0.4]{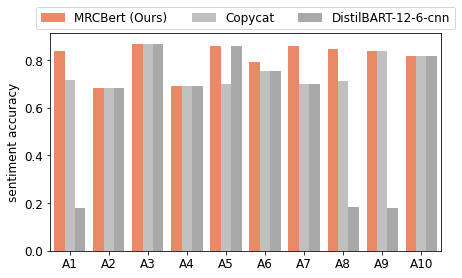}
    \includegraphics[scale=0.4]{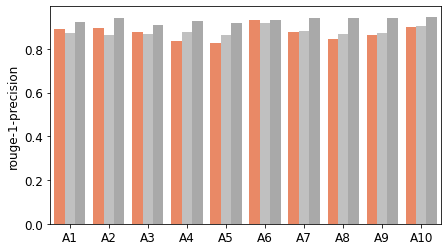}
    \includegraphics[scale=0.4]{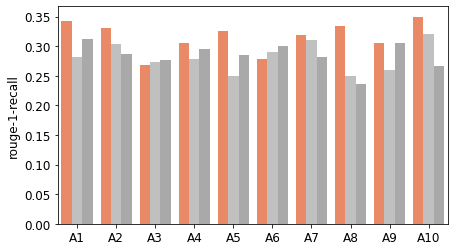}
    \includegraphics[scale=0.4]{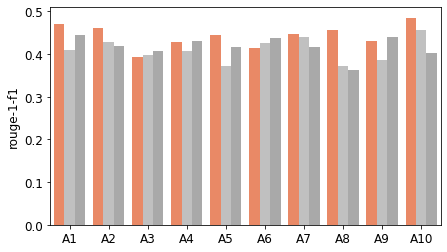}
    \caption{Sentiment accuracy and ROUGE-1 precision, recall and F1 scores for aspect-wise summaries, comparing MRCBert against Copycat and DistilBART-12-6-cnn. A1-A10 indicate the 10 aspects elaborated in Section \ref{section: proposed method, opinion}.}
    \label{fig: aspect wise sent and r1}
\end{center}
\vspace{-1.5em}
\end{figure}

Results in Figure \ref{fig: aspect wise sent and r1} show that MRCBert captures sentiment better than Copycat and DistilBART-12-6-cnn separately. The generated summaries also used relevant words from the original reviews as seen from the high precision. While the Recall is low, the low occurrence of 1-grams from the input reviews present in the generated summaries is intuitive as the input reviews are usually lengthy and may contain irrelevant information or incorrect use of words.

\subsection{Overall Results}
In table \ref{tab:aggrevated-result}, we show the sentiment score, ROUGE-1 and ROUGE-2 score for Copycat, DistilBART-12-6-cnn and our model- MRCBert, these results are obtained by averaging each model's performance. 

% \subsubsection{With additional pre-processing}
% Please add the following required packages to your document preamble:
% \usepackage{booktabs}
\begin{table}[htb]
\centering
\caption{Overall results with pre-processing}
\begin{tabular}{@{}lccccccc@{}}

\toprule
                    & \multicolumn{1}{l}{} & \multicolumn{3}{l}{ROUGE-1}     & \multicolumn{3}{l}{ROUGE-2}     \\ \midrule
                    & $S_{sentiment}$          & F1    & Precision      & Recall & F1    & Precision      & Recall \\
Copycat             & 0.675                & 0.479 & \textbf{0.992} & 0.331  & 0.132 & \textbf{0.868} & 0.074  \\
DistilBART-12-6-cnn & 0.687                & 0.526 & 0.990          & 0.378  & 0.148 & 0.717          & 0.087  \\
MRCBert & \textbf{0.798} & \textbf{0.533} & 0.980 & \textbf{0.386} & \textbf{0.162} & 0.630 & \textbf{0.099} \\ \bottomrule
\end{tabular}
\label{tab:aggrevated-result}
\end{table}

\subsection{Qualitative Analysis}
\label{ref: qualitative analysis}
In this section, we present some generated summaries for qualitative analysis. A summary about the display of an iPad after giving using MRCBert: \emph{"Product arrived on time and in good condition, works great, easy to use, good graphics. Screen is a bit darker than I expected it to be but it works great. Screen protector is not clear on the screen, but screen resolution is not as clear as the screen resolution. Sim card is a good size, but not compatible with sim card."}. 

While the summary generated is linguistically coherent and talks about the display of the iPad as designed, the last sentence about the sim card is not logical. We have yet to find a better evaluation metric or methodology that can capture logical fallacies of our generated summaries, and research that can flag logical fallacies in language modeling will be beneficial to consider as future work.

We also looked into rating-wise reviews to look for contrasting opinions and as expected, the rating 5 summary captured positive sentiment, unlike the rating 1 summary. An example rating 1 summary is as follows: \emph{The screen is black. Damaged without proper cable. It freezes all the time. It can't even open Safari. Not worth the purchase. I'm returning it.. out garageband app. It was a fraud. Cannot figure out how to switch screens cannot operate it at all.}

Meanwhile, an example rating 5 summary: \emph{Product is awesome, excellent seller, great product. For his birthday and is very happy with its operation. iPad looks new, works well and came in great condition to be used. Packed safely, worked like new. Fast shipping! Great product - works perfect. Arrived about a week early \& I am VERY glad I ordered it.}

\section{Future Works}
We mentioned earlier we would like to look into a summary generation that enables both linguistic and logical coherence as future work. Besides that, as we have demonstrated MRCBert for only electronics would like to extend it to include a larger range of products with varying attributes. Currently, our components are fragmented as we need to process BERT/DistilBERT and summary generation modules separately. The vision next is to adopt end-to-end modeling. Apart from these, the current work does not differentiate between genuine and fake reviews and generates summaries from all the reviews uniformly. Looking at generating summaries from only genuine reviews can be a potential future research topic. One possible idea is to weigh summaries according to some credibility score of reviewers. In our evaluation of the summaries, we used ROUGE-1, ROUGE-2, sentiment analysis, and qualitative analysis, however the set of metrics that we used is by no mean the most complete set of metrics that evaluate summaries to full extent, as for future work, a better set of metrics can be developed to evaluate summaries, this will guide researchers in developing summarization models in future. Lastly, our work thus far studies only mono-lingual reviews, but we can also look to summarize multi-lingual reviews.

% \newpage \clearpage
\bibliographystyle{plainnat}
\bibliography{main}

\newpage \clearpage
\section{Appendix}
% \subsection{ROUGE-2 scores for aspect-wise summaries, comparing MRCBert, Copycat, and DistilBART-12-6-cnn}
\begin{figure}[!htb]
\begin{center}
    \includegraphics[scale=0.4]{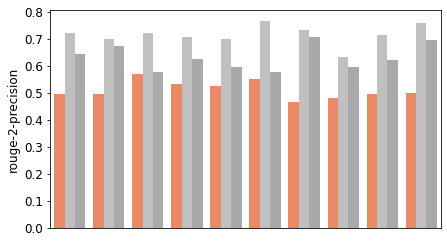}
    \includegraphics[scale=0.4]{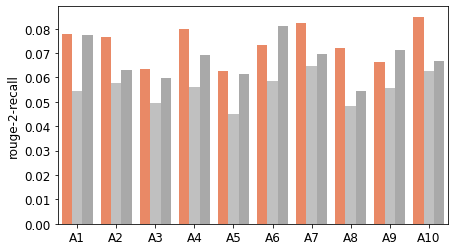}
    \includegraphics[scale=0.4]{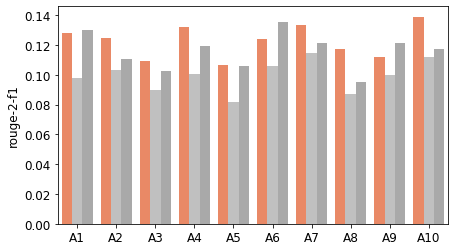}
    \caption{ROUGE-2 scores for aspect-wise summaries, comparing MRCBert, Copycat, and DistilBART-12-6-cnn.}
\end{center}
\label{fig: rouge_aspect_wise}
\end{figure}

% \subsection{Question Variants}
\begin{table}[!htb]
\centering
\caption{Question Variants}
\begin{tabular}{p{0.3\linewidth} p{0.3\linewidth} p{0.4\linewidth}}
\toprule
Feature          & Variant 1                & Variant 2                            \\ \midrule
Display          & How is display?          & What is opinion on display?          \\
Memory           & How is memory?           & What is opinion on memory?           \\
Speaker          & How is speaker?          & What is opinion on speaker?          \\
Sound            & How is sound?            & What is opinion on sound?            \\
Processor        & How is processor?        & What is opinion on processor?        \\
WiFi             & How is WiFi?             & What is opinion on WiFi?             \\
Battery          & How is battery?          & What is opinion on battery?          \\
Brand            & How is brand?            & What is opinion on brand?            \\
Operating System & How is operating system? & What is opinion on operating system? \\
Camera           & How is camera?           & What is opinion on camera?           \\ \bottomrule
\end{tabular}
% \caption{Question Variants}
\label{table:question}
\end{table}

% \subsection{Sample Summaries}
\begin{table}[!htb]
\centering
\caption{Sample summaries generated from MRCBert using opinion extraction}
\begin{tabular}{p{0.3\linewidth} | p{0.7\linewidth}}
\toprule
Feature &
  Summary \\ \midrule
Display &
  Product arrived on time and in good condition. Gps unit works great, but it does not work very well on my ipad. Keyboard is slow, but not the best of all of the features I have. Best gps device I have ever owned. Great value for the money you pay for. \\
Wireless Connection &
  Sound quality of this computer is excellent, but I wish it had. works great for the price I would recommend it to anyone who want to. product arrived on time and in good condition. Screen protector is a great value for the price .. however, I can't. was not able to return the screen protector, but it did not work. \\
Battery &
  This is an excellent product for the price I paid for it and the price. The display is nice, but not as good as the original one I have. The sound quality of this tablet is not bad, but not worth the money. This is a great little device to use with use to be able to clean up and clean up.\\
Operating System &
  The sound quality of these speakers are great for the price you can't go wrong. The ipad touch screen is great but the screen resolution is not clear enough to be \textless{}UNK\textgreater{}. The speakers on the ipad stand up to the touch screen but it can be not compatible with the iPad touch screen. \\
Processor &
  This is a great tablet for the price you can't go wrong it will be the best. The screen is a bit darker than I thought it would be but otherwise it. works well for what I need it to do, but have not been able to connect to my \textless{}UNK\textgreater page spread.\\
Brand &
  The screen is a great touch screen protector but the screen doesn't stay lit up. The ipad is not compatible with the ipad 2. The screen protector arrived on time and in good condition and the price was excellent. I would recommend this to anyone looking for a good price and a good deal of easy to use. \\
Memory &
  I have used this product for several months now and it has been holding up well. The cable is a great value for the price .. I have not had any problems. The only thing I dislike about is not to be \textless{}UNK\textgreater is not the best ipad. \\
Sound &
  The sensitivity of the sensitivity is not very what else can be said to say about a sensitivity machine. The product arrived on time and in good condition as advertised. The best tablet for the price you can't go wrong \textless{}UNK\textgreater the best ipad I have ever owned. \\
Camera &
  The screen protector is a great addition to the touch screen, but battery life is a bit long but it does not take up much space on. The product arrived in a timely manner and packaged well packed with no damage to. the product arrived on time and in good condition.. I can't wait to use it on my ipad 2 or 3 times a month. \\
Quality of Speaker &
  Reef is a great product for the price you can't go wrong with the \textless{}UNK\textgreater{}. it does what it says it will do, however, I wish it had a little. good power cord for the price I would recommend it to anyone looking for a good case cover for the price. this is the best ipad I have ever owned. \\ \bottomrule
\end{tabular}
% \caption{Sample summaries generated from MRCBert using Opinion Extraction.}
\label{tab:all-summaries-table}
\end{table}

% \subsection{All Results}
\newpage \clearpage
\begin{longtable}[c]{p{0.3\linewidth}  p{0.05\linewidth}  p{0.15\linewidth} p{0.05\linewidth} p{0.05\linewidth} p{0.05\linewidth} p{0.05\linewidth} p{0.05\linewidth} p{0.05\linewidth} p{0.05\linewidth}}
\caption{All results from Copycat, DistilBART-12-6-cnn and MRCBert using opinion extraction and without}\\
\toprule
  &                         &                      &           & \multicolumn{3}{c}{ROUGE-1} & \multicolumn{3}{c}{ROUGE-2} \\* 
\endhead
\bottomrule
\endfoot
\endlastfoot
Type                          & OE & model                & S & F1     & P & R & F1     & P & R \\ \midrule
all-reviews                   & no                      & Copycat               & 0.707     & 0.718  & 0.801     & 0.698  & 0.409  & 0.564     & 0.398  \\
rating1                       & no                      & Copycat               & 1.000     & 0.728  & 0.780     & 0.727  & 0.461  & 0.654     & 0.421  \\
rating2                       & no                      & Copycat               & 0.226     & 0.000  & 0.000     & 0.000  & 0.000  & 0.000     & 0.000  \\
rating3                       & no                      & Copycat               & 0.613     & 0.740  & 0.751     & 0.774  & 0.451  & 0.500     & 0.500  \\
rating4                       & no                      & Copycat               & 1.000     & 0.687  & 0.805     & 0.646  & 0.416  & 0.621     & 0.391  \\
rating5                       & no                      & Copycat               & 1.000     & 0.693  & 0.818     & 0.643  & 0.404  & 0.681     & 0.340  \\
how\_is\_display              & yes                     & Copycat               & 0.716     & 0.409  & 0.872     & 0.282  & 0.098  & 0.722     & 0.054  \\
how\_is\_wireless\_connection & yes                     & Copycat               & 0.685     & 0.430  & 0.863     & 0.303  & 0.103  & 0.701     & 0.058  \\
how\_is\_battery              & yes                     & Copycat               & 0.869     & 0.399  & 0.869     & 0.273  & 0.090  & 0.724     & 0.050  \\
how\_is\_operating\_system    & yes                     & Copycat               & 0.693     & 0.406  & 0.875     & 0.278  & 0.101  & 0.707     & 0.056  \\
how\_is\_processor            & yes                     & Copycat               & 0.698     & 0.373  & 0.862     & 0.250  & 0.082  & 0.700     & 0.045  \\
how\_is\_brand                & yes                     & Copycat               & 0.754     & 0.426  & 0.917     & 0.290  & 0.106  & 0.768     & 0.059  \\
how\_is\_memory               & yes                     & Copycat               & 0.700     & 0.440  & 0.880     & 0.310  & 0.115  & 0.735     & 0.065  \\
how\_is\_sound                & yes                     & Copycat               & 0.711     & 0.372  & 0.867     & 0.250  & 0.087  & 0.632     & 0.048  \\
how\_is\_camera               & yes                     & Copycat               & 0.839     & 0.385  & 0.874     & 0.260  & 0.100  & 0.716     & 0.056  \\
how\_is\_quality\_of\_speaker & yes                     & Copycat               & 0.819     & 0.456  & 0.905     & 0.320  & 0.112  & 0.761     & 0.063  \\
rating1                       & yes                     & Copycat               & 1.000     & 0.433  & 0.871     & 0.303  & 0.096  & 0.761     & 0.053  \\
rating2                       & yes                     & Copycat               & 0.226     & 0.426  & 0.905     & 0.292  & 0.102  & 0.778     & 0.056  \\
rating3                       & yes                     & Copycat               & 0.387     & 0.422  & 0.903     & 0.290  & 0.098  & 0.799     & 0.054  \\
rating4                       & yes                     & Copycat               & 1.000     & 0.405  & 0.893     & 0.274  & 0.101  & 0.772     & 0.056  \\
rating5                       & yes                     & Copycat               & 0.613     & 0.420  & 0.883     & 0.289  & 0.103  & 0.773     & 0.057  \\
all-reviews                   & no                      & DistilBART-12-6-cnn & 0.707     & 0.665  & 0.886     & 0.565  & 0.381  & 0.580     & 0.343  \\
rating1                       & no                      & DistilBART-12-6-cnn & 1.000     & 0.782  & 0.823     & 0.795  & 0.440  & 0.571     & 0.429  \\
rating2                       & no                      & DistilBART-12-6-cnn & 1.000     & 0.745  & 0.804     & 0.740  & 0.453  & 0.523     & 0.495  \\
rating3                       & no                      & DistilBART-12-6-cnn & 0.387     & 0.710  & 0.849     & 0.639  & 0.455  & 0.546     & 0.464  \\
rating4                       & no                      & DistilBART-12-6-cnn & 0.613     & 0.690  & 0.879     & 0.610  & 0.402  & 0.661     & 0.354  \\
rating5                       & no                      & DistilBART-12-6-cnn & 0.613     & 0.719  & 0.872     & 0.653  & 0.395  & 0.631     & 0.344  \\
how\_is\_display              & yes                     & DistilBART-12-6-cnn & 0.181     & 0.445  & 0.924     & 0.311  & 0.130  & 0.645     & 0.077  \\
how\_is\_wireless\_connection & yes                     & DistilBART-12-6-cnn & 0.685     & 0.420  & 0.943     & 0.286  & 0.111  & 0.673     & 0.063  \\
how\_is\_battery              & yes                     & DistilBART-12-6-cnn & 0.869     & 0.407  & 0.909     & 0.277  & 0.103  & 0.576     & 0.060  \\
how\_is\_operating\_system    & yes                     & DistilBART-12-6-cnn & 0.693     & 0.430  & 0.925     & 0.296  & 0.119  & 0.627     & 0.069  \\
how\_is\_processor            & yes                     & DistilBART-12-6-cnn & 0.861     & 0.416  & 0.919     & 0.285  & 0.106  & 0.595     & 0.061  \\
how\_is\_brand                & yes                     & DistilBART-12-6-cnn & 0.754     & 0.439  & 0.931     & 0.301  & 0.136  & 0.579     & 0.081  \\
how\_is\_memory               & yes                     & DistilBART-12-6-cnn & 0.700     & 0.417  & 0.942     & 0.282  & 0.121  & 0.706     & 0.070  \\
how\_is\_sound                & yes                     & DistilBART-12-6-cnn & 0.183     & 0.363  & 0.941     & 0.237  & 0.095  & 0.595     & 0.055  \\
how\_is\_camera               & yes                     & DistilBART-12-6-cnn & 0.181     & 0.440  & 0.939     & 0.305  & 0.121  & 0.623     & 0.071  \\
how\_is\_quality\_of\_speaker & yes                     & DistilBART-12-6-cnn & 0.819     & 0.402  & 0.947     & 0.266  & 0.117  & 0.696     & 0.067  \\
rating1                       & yes                     & DistilBART-12-6-cnn & 1.000     & 0.459  & 0.872     & 0.329  & 0.129  & 0.516     & 0.077  \\
rating2                       & yes                     & DistilBART-12-6-cnn & 0.613     & 0.479  & 0.920     & 0.342  & 0.135  & 0.606     & 0.080  \\
rating3                       & yes                     & DistilBART-12-6-cnn & 0.387     & 0.391  & 0.950     & 0.258  & 0.112  & 0.680     & 0.064  \\
rating4                       & yes                     & DistilBART-12-6-cnn & 0.613     & 0.436  & 0.908     & 0.301  & 0.124  & 0.637     & 0.072  \\
rating5                       & yes                     & DistilBART-12-6-cnn & 1.000     & 0.430  & 0.936     & 0.293  & 0.128  & 0.686     & 0.074  \\
how\_is\_display              & yes                     & MRCBert              & 0.839     & 0.471  & 0.890     & 0.342  & 0.128  & 0.496     & 0.078  \\
how\_is\_wireless\_connection & yes                     & MRCBert              & 0.685     & 0.460  & 0.896     & 0.330  & 0.125  & 0.497     & 0.076  \\
how\_is\_battery              & yes                     & MRCBert              & 0.869     & 0.393  & 0.875     & 0.267  & 0.109  & 0.569     & 0.064  \\
how\_is\_operating\_system    & yes                     & MRCBert              & 0.693     & 0.429  & 0.834     & 0.306  & 0.132  & 0.534     & 0.080  \\
how\_is\_processor            & yes                     & MRCBert              & 0.861     & 0.446  & 0.829     & 0.325  & 0.106  & 0.525     & 0.063  \\
how\_is\_brand                & yes                     & MRCBert              & 0.794     & 0.415  & 0.930     & 0.279  & 0.124  & 0.550     & 0.073  \\
how\_is\_memory               & yes                     & MRCBert              & 0.859     & 0.448  & 0.877     & 0.318  & 0.134  & 0.467     & 0.082  \\
how\_is\_sound                & yes                     & MRCBert              & 0.845     & 0.456  & 0.845     & 0.334  & 0.117  & 0.481     & 0.072  \\
how\_is\_camera               & yes                     & MRCBert              & 0.839     & 0.431  & 0.861     & 0.305  & 0.112  & 0.496     & 0.066  \\
how\_is\_quality\_of\_speaker & yes                     & MRCBert              & 0.819     & 0.485  & 0.900     & 0.350  & 0.139  & 0.502     & 0.085  \\* \bottomrule
\label{tab:all-results-table}\\
\end{longtable}

% \newpage \clearpage
% \bibliographystyle{plainnat}
% \bibliography{main}
\end{document}